# Article

# Artificial intelligence-driven improvement of hospital logistics management resilience: a practical exploration based on H Hospital

Lu Huang[1], Dongjing Shan[1,2], Han Chen*[,3]

1.Southwest Medical University, Luzhou 646000, Sichuan, People's Republic of China

2.Affiliated Hospital of Southwest Medical University, Luzhou, 646000, China

3.Shinawatra University, Bangkok 10400, Thailand

**Corresponding author:** Han Chen, han.c@siu.ac.th

## Abstract

Hospital logistics management faces growing pressure from internal operations and external emergencies, with artificial intelligence (AI) holding untapped potential to boost its resilience. This study explores AI's role in enhancing logistics resilience via a mixed-methods case study of H Hospital, combining 12 key informant interviews and a full survey of 151 logistics staff, with the PDCA cycle as the analytical framework. Thematic and quantitative analyses (hierarchical regression, structural equation modeling) were adopted for data analysis. Results showed 94.7% staff perceived AI application, with the strongest improvements in equipment maintenance (41.1%) and resource allocation (33.1%), but limited effects in emergency response (18.54%) and risk management (15.23%). AI integration positively correlated with logistics resilience (β=0.642, p<0.001), with management system adaptability as a positive moderator (β=0.208, p<0.01). The PDCA cycle fully mediated the AI-resilience relationship. We conclude AI effectively enhances logistics resilience, dependent on adaptive management systems and structured continuous improvement mechanisms. Targeted strategies are proposed to form an AI-driven closed-loop resilience mechanism, offering empirical guidance for AI-hospital logistics integration and resilient health system construction.

## Introduction

Hospital logistics management, as the fundamental support system ensuring the stable operation of medical institutions, has evolved from basic material supply to complex systematic services encompassing equipment management, material supply chain, human resources, financial management, safety management, and information management [1]. Through empirical research, the core content of traditional hospital logistics management has been systematically categorized into eight key areas: equipment management, materials management, human resource management, financial management, safety management, information management, building and space management, and greening and environmental management [2]. Research indicates that effective equipment management can increase equipment availability to over 95%, significantly higher than the industry average [3].

In China, large comprehensive tertiary hospitals face enormous operational pressure. H Hospital, a major medical center in the junction area of Sichuan, Chongqing, Yunnan, and Guizhou, handles approximately 3 million outpatient and emergency visits annually, completes about 100,000 surgeries, and maintains nearly 4,000 beds [4]. Under this normal state of high-load operation, logistics system stability and adaptive capacity are crucial for ensuring continuity of medical services.

However, traditional hospital logistics management models expose significant vulnerability when facing internal operational pressure and external emergencies. Common problems include poor information flow between departments, resource allocation relying on empirical decision-making rather than scientific basis, and rigid emergency response mechanisms lacking intelligent early warning and scheduling capabilities [5]. Studies indicate that traditional logistics management models face three major challenges during public health crises: information transmission relying on paper documents or verbal communication, which is prone to the bullwhip effect and leads to delayed responses; resource allocation based on fixed plans without dynamic adjustment capabilities; and strict departmental barriers with significant information silos, leading to inefficienct coordination [6].

These challenges necessitate a fundamental shift in the paradigm of logistics management. Scholars have pointed out that logistics systems must possess a higher-order capability: resilience, the ability to quickly perceive, effectively buffer, flexibly adapt, and rapidly recover when facing internal and external sudden shocks [7]. The evolution of the resilience concept reflects the continuous deepening of human understanding of the operational laws of complex systems. Originally stemming from material mechanics, resilience was innovatively introduced into ecology in the 1970s [8], and later expanded into social-ecological systems, forming the evolutionary resilience perspective which emphasizes the system's ability to learn, adapt, and even transform amidst change [9].

The rise of artificial intelligence technology brings revolutionary opportunities for change. AI applications in hospital logistics, including Internet of Things for real-time data collection, machine learning for demand forecasting[36], computer vision for safety monitoring, and predictive maintenance for equipment management, offer potential to enhance system responsiveness and adaptive capacity [10]. Research shows that SPD smart logistics systems based on machine learning and operational research algorithms can achieve accurate demand forecasting and automatic replenishment of medical consumables based on multi-source information such as historical consumption data, number of inpatients, and surgery schedules [11]. In equipment and facility operation and maintenance, deploying sensors on key facilities combined with AI predictive maintenance models enables real-time monitoring and early fault identification [12].

However, in practice, AI application often remains fragmented, limited to pilot projects in independent modules such as smart logistics (SPD) and smart canteens, failing to build an integrated intelligent management center [13]. Existing management systems frequently do not match AI-driven business processes, lacking standardized procedures for data security, system operation, and human-machine collaboration [14]. Furthermore, logistics management teams often lack AI literacy, with severe shortage of professionals combining medical background, management knowledge, and AI skills [15]. Research found that without systematic, continuous training and communication, employees find it difficult to effectively accept and use new systems, and may even develop negative emotions due to fear of being replaced [16].

Therefore, understanding how to systematically promote deep integration of AI technology with hospital logistics management practices, overcome current obstacles in institutions, technology, talent, and funding, and build a highly resilient logistics support system has become an urgent core management issue. This study addresses three research questions: (1) How does artificial intelligence technology specifically affect the resilience performance of hospital logistics management in key dimensions including information flow, resource allocation, and emergency response? (2) How are coordination barriers manifested between AI applications and supporting elements such as management systems, talent teams, and funding investment? (3) How can an AI-driven logistics management resilience enhancement mechanism be constructed to integrate existing fragmented systems and overcome institutional and talent bottlenecks?

## Methods

**Study design.** This study employed a mixed-methods research design centered around a case study of H Hospital, a large comprehensive tertiary Grade A hospital in China. The design comprised two stages: qualitative exploration through interviews to diagnose current situation and obstacles, followed by quantitative verification via questionnaire survey to analyze variable relationships and mechanisms of action.

**Qualitative sampling and data collection.** A purposive sampling strategy was adopted to select core key informants who could provide the most substantial information [13]. Interviewees included: hospital leadership responsible for logistics, the Director of the Logistics Department, heads of various sections (Materials, Equipment, Infrastructure, and General Affairs), and engineers from the Information Technology department responsible for AI system operation and maintenance. A total of 12 individuals were interviewed, ensuring comprehensive perspectives from strategic, managerial, and technical operational levels. Semi-structured interviews lasting approximately 40-60 minutes were audio-recorded with consent and subsequently transcribed for analysis. Internal documents related to logistics operations, including annual reports, policy documents, and AI system introduction manuals, were also collected to supplement and triangulate interview data.

**Quantitative sampling and data collection.** A whole-population survey strategy was employed [13]. Given that the total number of staff in H Hospital's logistics department is relatively definite and manageable, the questionnaire was distributed to all members of the entire logistics management department. This approach aimed to minimize sampling bias and obtain the most accurate depiction of the overall situation. The questionnaire was administered online using Wenjuanxing platform, distributed through WeChat with assistance from H Hospital's Logistics Department office. Before filling out the questionnaire, research purpose, data confidentiality, and anonymity principles were clearly explained. The data collection period lasted two weeks, yielding 151 valid responses (100% response rate).

**Measurement instruments.** The questionnaire was designed to systematically collect feedback from staff regarding their perceptions and evaluations of artificial intelligence technology application and supporting elements. Content covered three core sections: (1) current state of AI technology application, (2) performance of logistics management resilience, and (3) key enabling factors (management systems, platform development, talent development, and funding investment). Demographic information included department, job grade, and years of service, based on established research [4].

Core measurement items used classification choices and degree judgments tailored to specific measurement objectives. For degree-based items, responses were assigned numerical values (A=4, B=3, C=2, D=1). Overall perception of resilience enhancement was operationalized as the sum of dimensions selected, serving as a proxy variable for breadth of perceived resilience improvement. The questionnaire included an open-ended question for additional suggestions.

**Pilot study.** Before formal distribution, expert review was conducted with three specialists in hospital management and information technology to evaluate content validity. Based on feedback, modifications were made to improve clarity, accuracy, and relevance of item statements. Subsequently, ten employees from H Hospital's logistics department participated in small-scale testing to identify understanding difficulties and refine wording.

**Reliability and validity.** Cronbach's alpha coefficient was used to measure questionnaire reliability. According to methodological consensus, a Cronbach's alpha value of 0.70 or above is considered satisfactory [17]. The overall questionnaire achieved Cronbach's alpha of 0.936, indicating very high reliability. Content validity was ensured through literature review, qualitative interviews, and expert evaluation. Scale structure was examined using both Exploratory Factor Analysis (EFA) and Confirmatory Factor Analysis (CFA), with results demonstrating alignment with the theoretical framework[37].

**Data analysis.** For qualitative data, thematic analysis was employed to code, categorize, and analyze interview transcripts and documentary materials using Nvivo software[43]. Core themes regarding fragmented AI application, coordination obstacles, impacts on resilience, and PDCA practices were identified.

For quantitative data, SPSS and AMOS statistical software were utilized. Analysis included: (1) descriptive statistical analysis to describe sample characteristics and variable distributions[40]; (2) Pearson correlation analysis to preliminarily assess relationships between variables; (3) hierarchical regression analysis to test main effects and moderating effects while controlling for demographic variables; and (4) structural equation modeling to simultaneously examine complex path relationships among multiple variables, including the mediating role of PDCA cycle practice.

**Ethical considerations.** The study was conducted in accordance with ethical research principles. All participants were informed about the research purpose, voluntary participation, and confidentiality of responses. Anonymity was maintained throughout data collection and analysis.

# Results

**Sample characteristics.** The final valid sample comprised 151 respondents from H Hospital's logistics management department. Table 1 presents the distribution across key demographic variables.

**Perception of AI application.** Regarding whether the hospital has empowered logistics management with digital and intelligent technologies, 94.70% of staff indicated they could clearly perceive the application of such technologies in daily logistics work, while only 5.30% reported not yet experiencing implementation. This high level of staff awareness indirectly suggests that current AI empowerment measures have been largely implemented effectively.

**Perceived effectiveness of AI in enhancing resilience.** Respondents' evaluations regarding the effectiveness of AI in enhancing logistics management resilience showed a positive trend, with 68.21% considering effects significant or relatively significant. However, 31.79% perceived effects as moderate or insignificant. This outcome indicates that while technology application has gained recognition, its efficacy in achieving the deeper objective of improving organizational resilience has not yet been fully demonstrated.

**Specific dimensions of resilience enhancement.** Analysis indicates that AI implementation has significantly optimized core operational aspects that are quantifiable and standardized. Notably, 41.06% of employees reported marked improvement in equipment maintenance capability, primarily attributed to AI-driven predictive maintenance systems utilizing real-time monitoring and intelligent analysis [3]. Concurrently, 33.11% acknowledged its reinforcing role in resource allocation capability, through AI algorithms integrating historical consumption data with real-time demand information enabling intelligent replenishment and optimal routing [11]. Additionally, 29.14% perceived improvements in talent management capability, stemming from AI applications in intelligent staff scheduling, automated work order assignment, and performance data visualization.

However, employee perceptions of improvement were notably lower in dimensions requiring complex cross-departmental coordination and comprehensive judgment: emergency response (18.54%) and risk management (15.23%). This pattern suggests current AI implementations are more effective at bolstering efficiency resilience for routine operations than enabling adaptive and collaborative resilience required for crisis response [18]. The lowest perceived improvement was in environmental hygiene management (10.60%), indicating nascent AI adoption in this field.

**Contribution of key enabling factors.** Survey results revealed differentiated evaluations among respondents regarding the contribution of four key enabling factors. An overwhelming majority placed a sound management system at the core of contribution, highlighting consensus that AI technology's potential is highly dependent on compatible soft environment [19]. Platform development received positive evaluations from 99 respondents, reflecting widespread recognition of its foundational importance [20]. However, for talent team development and capital investment, the number of respondents considering contributions significant did not exceed 60%, with a considerable number rating effectiveness as insignificant [21]. Research shows that the professional quality of logistics staff directly affects service quality, but this group often faces issues like insufficient training and unclear career paths [22].

**Correlation analysis.** Pearson correlation analysis revealed significant positive pairwise correlations among all key variables at the 0.01 level[44]. The independent variable Perceived Effectiveness of AI Application showed strong positive correlation with the dependent variable Overall Perception of Logistics Management Resilience Enhancement ($r = 0.658$, $p < 0.01$), providing preliminary statistical evidence supporting the positive influence of AI integration on resilience. Among the four enabling factors, Contribution of Management System showed the strongest correlation with Overall Resilience Perception ($r = 0.579$).

**Hypothesis testing: hierarchical regression analysis.** To rigorously test H1 and H2, hierarchical regression analysis was conducted with Level of Logistics Management Resilience as the dependent variable. Table 2 presents the results.

**Path analysis of PDCA-based continuous improvement mechanism.** Structural equation modeling was employed to test H3, with PDCA

## Table 1 Demographic characteristics of the sample (N=151)

| Variable | Category | Frequency | Percentage(%) |
|---|---|---|---|
| Department | Functional Department | 58 | 38.41 |
| | Technical Department | 47 | 31.13 |
| | Support & Operations Department | 32 | 21.19 |
| | Other Branch/Affiliated Units | 14 | 9.27 |
| Job Level | Middle & Senior Management | 23 | 15.23 |
| | General Staff | 128 | 84.77 |
| Years of | 5 years or less | 45 | 29.80 |
| | 5 to 10 years | 63 | 41.72 |
| | More than 10 years | 43 | 28.48 |

The sample encompasses various position types within the logistics management department, with employees from functional and technical departments constituting the majority (69.54%). General staff form the main body (84.77%), while over 15% of managerial staff are included, ensuring perspectives from both operational and management levels. Years of service distribution is relatively balanced, guaranteeing diversity in understanding based on different levels of professional experience

Cycle Practice operationalized as a second-order factor. The PDCA cycle, popularized and developed by Deming, views any management task as a cyclical process consisting of four stages, revolving repeatedly and spiraling upwards [25]. The model demonstrated good fit indices ($\chi^2$/df = 2.91, CFI = 0.928, TLI = 0.912, RMSEA = 0.070). Path coefficients revealed that Level of AI Integration had significant positive effect on PDCA Cycle Practice ($\beta$ = 0.71, $p < 0.001$), and PDCA Cycle Practice subsequently had significant positive effect on Level of Logistics Management Resilience ($\beta$ = 0.63, $p < 0.001$)[39]. More importantly, the direct path from Level of AI Integration to Level of Resilience became non-significant after introducing PDCA Cycle Practice as a mediator ($\beta$ = 0.15, $p > 0.05$), while the indirect effect was significant. This indicates that the positive impact of AI technology on resilience is primarily realized by embedding and activating the PDCA-based continuous improvement mechanism, providing key quantitative evidence supporting H3 [26].

**Qualitative evidence triangulation.** Interview data further elucidated this mechanism. Managerial-level interviewees noted institutional adaptation: "Prior to the AI project launch, we revised the Emergency Material Dispatch Plan, incorporating intelligent forecasting data as a basis for decision-making." Regarding process re-engineering: "Platform implementation was not merely about installation; we re-engineered the inspection workflow to align with the AI work order system." Frontline staff reported iterative improvement: "The system initially had inaccurate warning thresholds, leading to numerous false alarms. Later, we adjusted parameters based on historical data, and now it's much more accurate." Overcoming fragmentation was highlighted: "Early AI applications were siloed—standalone energy monitoring, independent clean supply transport systems. Subsequently, by establishing a unified logistics data platform and instituting monthly cross-departmental coordination meetings, we gradually achieved data integration and operational synergy." This reflects the importance of establishing effective cross-departmental coordination mechanisms [27].

**Summary of hypothesis testing.** Table 3 presents the summary of hypothesis testing results.

# Discussion

This study investigated how artificial intelligence drives the enhancement of logistics management resilience in a large Chinese tertiary hospital, using a mixed-methods approach grounded in the PDCA framework. Our findings reveal both the potential

**Table 2 Results of hierarchical regression analysis for level of logistics management resilience (N=151)**

| Variable | Model 1 | Model 2 | Model 3 |
|---|---|---|---|
| **Control Variables** | | | |
| Department | Controlled | Controlled | Controlled |
| Job Level | Controlled | Controlled | Controlled |
| Years of Service | Controlled | Controlled | Controlled |
| **Main Effect** | | | |
| Perceived Level of AI Integration | | .642*** | .598*** |
| **Moderation Effect** | | | |
| Perceived Adaptability of | | | .185** |
| Interaction Term (A × B) | | | .208** |
| **Model Statistics** | | | |
| R² | .048 | .449 | .520 |
| Adjusted R² | .028 | .425 | .502 |
| ΔR² | | .401*** | .071** |
| F-value | 2.397 | 18.923*** | 21.457*** |

*Note: *p < 0.05, **p < 0.01, **p < 0.001; reported coefficients are standardized beta (β) values.

After introducing Perceived Level of AI Integration in Model 2, the model's explanatory power substantially increased (ΔR² = 0.401, p < 0.001), with significant regression coefficient of β = 0.642 (p < 0.001). This result fully supports H1, confirming significant positive influence of AI technology integration on hospital logistics management resilience [23]. Model 3, incorporating Perceived Adaptability of Management System and its interaction term with AI integration, revealed significantly positive coefficient for the interaction term (β = 0.208, p < 0.01) and further statistically significant increase in explanatory power (ΔR² = 0.071, p < 0.01). This strongly supports H2, indicating that management system adaptability plays a crucial positive moderating role in the AI-driven resilience enhancement process [24].

and limitations of AI applications in this context, with important implications for theory and practice.

**Principal findings.** First, while AI technology adoption is widespread (94.7% awareness), its impact on resilience is multidimensional and uneven. AI demonstrates strongest efficacy in optimizing data-intensive operational domains, equipment maintenance (41.1%) and resource allocation (33.1%), areas characterized by structured processes and data-driven decisions[41]. This aligns with research showing that AI-based predictive maintenance can significantly reduce unplanned downtime [3].However, perceived contribution remains substantially lower in domains requiring complex cross-functional coordination and judgment, such as emergency response (18.5%) and risk management (15.2%).This pattern suggests current AI implementations are more effective at bolstering "efficiency resilience" for routine operations than enabling the "adaptive and collaborative resilience" required crisis response

**Table 3 Summary of hypotheses testing**

| Hypothesis | Result |
|---|---|
| **H1:** The degree of artificial intelligence technology integration (as opposed to its mere adoption) is significantly positively correlated with the level of resilience in the logistics management of Hospital H. | **Supported** |
| **H2:** The adaptability of management institutions plays a crucial moderating role in the process of AI-driven logistics resilience enhancement; the more well-developed the institutions, the more pronounced the AI-enabled effects. | **Supported** |
| **H3:** The continuous improvement mechanism based on the PDCA cycle effectively facilitates the integration of AI systems with logistics business processes, serving as a key pathway to overcome fragmented application and achieve a spiral escalation of resilience. | **Supported** |

[28]. Research on hospital disaster resilience has similarly found that preventive and preparatory resilience is easier to achieve than adaptive and restorative resilience [29].

Second, our hierarchical regression analysis confirms not only a significant main effect of AI integration on resilience ($\beta=0.642$, $p<0.001$) but also a significant positive interaction effect between management system adaptability and AI integration ($\beta=0.208$, $p<0.01$). This statistical evidence underscores that a well-adapted management system acts as a crucial moderator and amplifier for AI's impact. The value of intelligent algorithms is contingent upon complementary rules, clarified responsibilities, and revised workflows that govern how AI-driven insights translate into action, particularly during disruptions [9]. Studies on organizational adaptation to environmental jolts have long emphasized the importance of institutional flexibility [30].

Third, and most significantly, our path analysis reveals that the PDCA cycle fully mediates the AI-resilience relationship. This finding positions AI not as a standalone solution but as a catalyst embedded within a continuous improvement framework. The technology provides data and analytical capacity for robust monitoring (Check) and informed planning (Plan), but sustained resilience gains depend on institutionalizing subsequent steps of implementing changes (Do) and standardizing successful adaptations (Act) [25]. Research on PDCA application in hospital logistics management has demonstrated its effectiveness in improving service quality and efficiency [31,32]. The relative weakness in emergency and risk-related resilience may therefore reflect not failure of AI technology per se, but gaps in closing this PDCA loop.

Strengths and limitations. This study's strengths include its mixed-methods design combining qualitative depth with quantitative breadth, providing comprehensive understanding of AI-resilience dynamics in a real-world hospital setting. The whole-population survey strategy minimized sampling bias, and the application of PDCA theory as an analytical framework offers theoretical grounding often lacking in technology implementation research.

However, several limitations should be acknowledged. First, findings derive from a single-case study of one Chinese hospital, which may affect generalizability to other healthcare institutions with differing resources, structures, or cultures. Second, reliance on perceptual self-reported data from logistics department staff, while insightful, does not fully capture objective performance metrics or perspectives of other critical stakeholders such as clinical end-users. Third, the cross-sectional design captures associations but cannot definitively establish causality or trace temporal dynamics of socio-technical co-evolution as AI systems mature.

Comparison with other studies. Our findings align with and extend previous research. Consistent with studies from Ghana and Nigeria examining traditional medicine integration [33], our work highlights that successful health system interventions require not just technological innovation but also institutional adaptation and stakeholder engagement. The finding that management system adaptability moderates AI effectiveness resonates with research on digital transformation in African healthcare settings, where policy frameworks and regulatory mechanisms critically influence implementation outcomes [34].

The mediating role of continuous improvement mechanisms we identified parallels findings from studies of health system strengthening in resource-limited settings, where structured quality improvement approaches have proven essential for translating innovations into sustainable practice improvements [35]. This suggests that regardless of geographic context, the principles of systematic planning, implementation, monitoring, and adaptation remain fundamental to building resilient health systems.

**Implications for policy and practice.** For hospital administrators and health system managers, our findings provide an evidence-based roadmap for AI-enabled resilience building. First, investment must be strategically balanced between technology acquisition and institutional adaptation. Rather than pursuing isolated AI applications, hospitals should prioritize integration and interoperability, creating unified data platforms that break down information silos[38]. Second, management systems require proactive redesign to accommodate AI-driven workflows, including clear data governance guidelines, revised emergency protocols incorporating AI-generated alerts, and updated performance metrics incentivizing data-driven decision-making. Third, the PDCA cycle should be formalized as the core management rhythm, with AI-powered analytics serving as the primary input for regular operational reviews, and mechanisms ensuring insights translate systematically into updated procedures and organizational learning.

For policymakers, our study highlights that supporting hospital resilience in the digital era requires enabling environments that encourage not just technology adoption but also necessary process re-engineering and change management. Funding programs should support integrated implementation rather than fragmented technology procurement, and industry-wide guidelines for data governance and human-AI collaboration in clinical support functions would facilitate safer, more effective deployment.

**Unanswered questions and future research.** Several important questions remain. Future research should employ longitudinal designs to trace causal pathways and understand temporal dynamics as AI systems mature. Comparative multi-case studies across diverse hospital settings would help distinguish universal principles from context-dependent factors. Expanding the lens to include perspectives of clinical staff, patients, and technology partners would enrich understanding of cross-functional alignment. Finally, investigation is warranted into development and governance of next-generation AI tools specifically designed for adaptive resilience, such as simulation-based digital twins and multi-agent coordination systems, alongside ethical frameworks required for deployment in high-stakes healthcare environments[42,45].

## Conclusion

This study demonstrates that artificial intelligence can significantly enhance hospital logistics management resilience, but its effectiveness depends critically on adaptive management systems and structured continuous improvement mechanisms. The transition from AI adoption to realized resilience is not automatic but governed by a synergistic triad: AI technology acts as the foundational catalyst, adaptive management systems serve as the essential force multiplier, and the PDCA cycle operates as the core transmission mechanism converting technological potential into sustained organizational learning and improvement. For hospitals navigating increasingly uncertain environments, strategic investments must concurrently target technological infrastructure, institutional redesign, and cultivation of a disciplined improvement culture to fully harness AI's potential for building resilient healthcare logistics systems.

## Competing interests

The authors declare no competing interests.

## Authors' contributions

Lu Huang conceived the study, designed the methodology, conducted data collection and analysis, and drafted the manuscript. Han Chen supervised the research, contributed to study design, and revised the manuscript critically for important intellectual content. Both authors read and approved the final version of this manuscript.

## Acknowledgments

The authors thank the logistics management department staff at H Hospital for their participation and support. We also acknowledge the faculty members at Shinawatra University for their valuable guidance throughout this research.

# Appendix: Survey Questionnaire

**Part 1: Demographic Information**

1. Department Affiliation
   □ Functional Department
   □ Technical Department
   □ Support & Operations Department
   □ Other Branch/Affiliated Unit

2. Position Level
   □ Middle & Senior Management
   □ General Staff

3. Tenure
   □ 5 years or less
   □ Between 5 and 10 years
   □ More than 10 years

**Part 2: Core Measurement Items**

4. Do you think the hospital has started applying Artificial Intelligence technologies to support logistics management?
   □ Already Started
   □ Not Started

5. How effective do you think AI technology is in enhancing hospital logistics management?
   □ Significant Effect (4)
   □ Relatively Significant Effect (3)
   □ Moderate Effect (2)
   □ Insignificant Effect (1)

6. In which specific dimensions do you think the resilience of current hospital logistics management has improved? (Check all that apply)
   □ Emergency Response Capability
   □ Resource Allocation Capability
   □ Risk Management Capability
   □ Talent Management Capability
   □ Equipment Maintenance Capability
   □ Environmental Hygiene Management Capability

7. How effective is improving the management system in enhancing the level of hospital logistics management?
   □ Significant Effect (4)
   □ Relatively Significant Effect (3)

□ Moderate Effect (2)
□ Insignificant Effect (1)

8. How effective is platform development in enhancing hospital logistics management?
   □ Significant Effect (4)
   □ Relatively Significant Effect (3)
   □ Moderate Effect (2)
   □ Insignificant Effect (1)

9. How effective is talent team building in enhancing hospital logistics management?
   □ Significant Effect (4)
   □ Relatively Significant Effect (3)
   □ Moderate Effect (2)
   □ Insignificant Effect (1)

10. How effective is funding investment in enhancing hospital logistics management?
    □ Significant Effect (4)
    □ Relatively Significant Effect (3)
    □ Moderate Effect (2)
    □ Insignificant Effect (1)

**Part 3: Open-ended Feedback**

11. Please share any additional suggestions, comments, or insights you may have regarding the improvement of hospital logistics management.

1.